\documentclass[letterpaper, preprint, paper,11pt]{AAS}	

\usepackage{bm}
\usepackage{amsmath}
\usepackage{amsfonts}
\usepackage{subcaption}
\usepackage[colorlinks=true, pdfstartview=FitV, linkcolor=black, citecolor= black, urlcolor= black]{hyperref}
\usepackage{overcite}
\usepackage{footnpag}			
\usepackage{siunitx}

\PaperNumber{26-875}

\begin{document}

\title{PHYSICS-INFORMED CONDITIONAL NORMALIZING FLOWS FOR ANGLES-ONLY CISLUNAR ORBIT DETERMINATION}

\author{Walther Litteri\thanks{Ph.D. Student, Aerospace Centre of Excellence, University of Strathclyde, G11XQ Glasgow, UK},  
Massimiliano Vasile\thanks{Professor and Director, Aerospace Centre of Excellence, University of Strathclyde, G1 1XQ Glasgow, UK}
}

\maketitle{}

\begin{abstract}
Generative Astrodynamics is advanced in this work by extending generative modelling to an orbit determination problem in the cislunar environment. The task is formulated as conditional density estimation, aiming to infer the probability distribution of the initial state from angles-only measurements over short observation arcs. A normalising flow is trained on perturbed topocentric observations from Near Rectilinear Halo Orbits, enabling a flexible and potentially multimodal posterior representation. Given new measurements, the learned density is sampled to generate statistically consistent and physics-informed state hypotheses. These estimates are refined via nonlinear least-squares minimisation, providing a competitive warm start for classical algorithms.
\end{abstract}

\section{Introduction}
The problem of recovering the orbit followed by a body in space is arguably one of the first scientific problems ever formulated. With the advent of space exploration and the growing exploitation of the space environment, this problem has also become central to the engineering community, which requires reliable methods to estimate the state of a spacecraft from observational data. This field of research is known as \textit{Orbit Determination} (OD). The literature further distinguishes between different operational scenarios depending on the phase of the mission \cite{vallado_fundamentals_1997}. In particular, when orbit estimation is performed at the very beginning of operations, e.g., immediately after launch or following a critical recovery phase, the problem is referred to as Initial Orbit Determination (IOD). A key difference between standard OD and IOD lies in the availability of prior information. In routine OD operations, additional data may be available from previously estimated states or measurements provided by onboard instruments. When such information is unavailable, as in the case of newly launched spacecraft or uncooperative objects, the IOD problem becomes significantly more challenging.

Over the past decades, interest in space operations in the cislunar environment has increased considerably. The presence of water ice in permanently shadowed craters near the lunar south pole makes the Moon an attractive location for future scientific and commercial activities \cite{anand_lunar_2010}. The Artemis programme represents the most recent international effort aimed at returning humans to the lunar vicinity, with the crewed lunar flyby mission Artemis II now scheduled for April 2026 \cite{9843277}. In the original Artemis architecture, the Gateway lunar orbital outpost was conceived as a strategic counterpart to the Earth‑orbiting International Space Station, designed to support subsequent surface missions, including Artemis IV and V. However, following a comprehensive restructuring of the Artemis programme announced in early 2026, the role and priority of the Gateway have become significantly less certain, with current plans shifting towards a more direct and sustained human presence on the lunar surface.

Operating in the cislunar environment introduces several additional challenges compared to Earth-orbiting missions. The large Earth–Moon distance complicates communication and synchronisation, introducing significant time delays between command transmission and reception. Moreover, the intrinsic dynamical instability and chaotic behaviour that characterise many cislunar trajectories further increase operational complexity, reducing the margin for allowable errors and demanding higher levels of precision and accuracy. Orbit Determination in cislunar space is further complicated by the absence of a dedicated Global Navigation Satellite System (GNSS) capable of supporting navigation in the region, as well as by limited communication opportunities that restrict the availability of frequent tracking measurements.

Traditional approaches to orbit determination typically rely on correction schemes, such as minimising the nonlinear mean squared error in a statistical framework. The Extended Kalman Filter (EKF) integrates knowledge of the dynamics to improve estimation accuracy. More recently, machine learning and statistical learning techniques have been explored to model unmodelled forces through nonlinear latent force models, which act as filters for uncorrelated Gaussian noise \cite{hartikainen_state-space_2012, caldas_machine_2024}. Physics-Informed Neural Networks (PINNs) are neural networks trained with an additional term enforcing the underlying physical laws \cite{wang_experts_2023, watson_machine_2025}. In the context of OD, PINNs have been used to approximate orbit propagation and measurement models in the cislunar environment, thereby facilitating the solution of the inverse problem\cite{scorsoglio_physics-informed_2023}.

Generative Artificial Intelligence (AI) refers to deep learning strategies used to model probability distributions and to sample from them. Applications of this framework span a variety of domains, from the generation of text, images, and audio \cite{manduchi_challenges_2025}. In astrodynamics, within the emerging field of Generative Astrodynamics, generative methods based on Variational Autoencoders have been successfully applied to analyse and generate periodic orbits in the Circular-Restricted Three-Body Problem (CR3BP)\cite{litteri_generative_2025, litteri_generation_2026}, while Normalizing Flow models have been employed for the identification and characterisation of equilibrium solutions in the N-body problem\cite{wilson_generating_2026}.

This work presents a novel application of Generative Astrodynamics to the solution of an orbit determination problem in the cislunar environment. Topocentric states corresponding to an observation arc are modelled as a distribution conditioned on angles-only observations. A Normalizing Flow model is trained on this distribution, and sampling conditioned on observations provides the estimated states, which are further refined using traditional methods. It will be shown how the inclusion of an additional term in the loss, based on the propagation of the trajectory, helps regularise the physics loss, yielding more accurate results.

The paper is structured as follows. First, an overview of the problem is provided, defining the quantities of interest and the underlying assumptions. The model definition follows, detailing architectural choices, trade-off, and the implementation of the physics-informed loss. Finally, the performance of the proposed approach is discussed, including a comparison with traditional methods, and concluding remarks are presented.

\section{Cislunar Orbit Determination}
\subsection{Scenario}
\begin{figure}[h!]
    \centering
    \begin{subfigure}{0.45\linewidth}
        \centering
        \includegraphics[width = \linewidth]{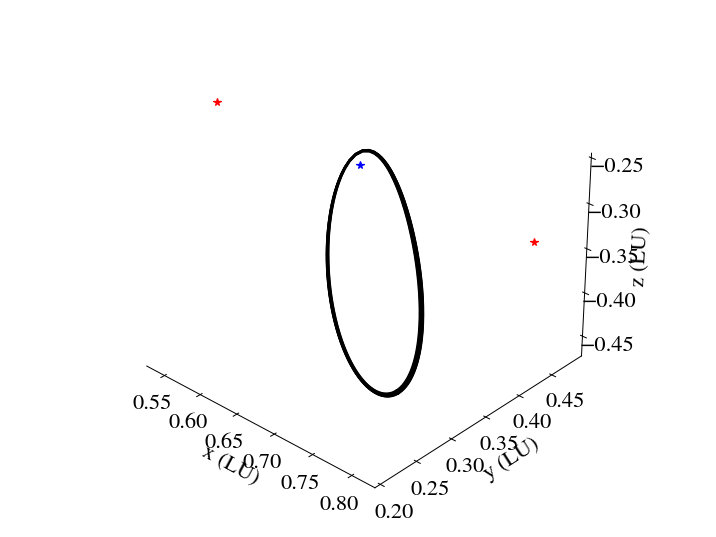}
        \caption{Synodic reference frame. The red stars denote the Euler-Lagrange points $L_1, L_2$ and the blue star the Moon}
        \label{fig:nrho_syn}
    \end{subfigure}
    \quad
    \begin{subfigure}{0.45 \linewidth}
        \centering
        \includegraphics[width = \linewidth]{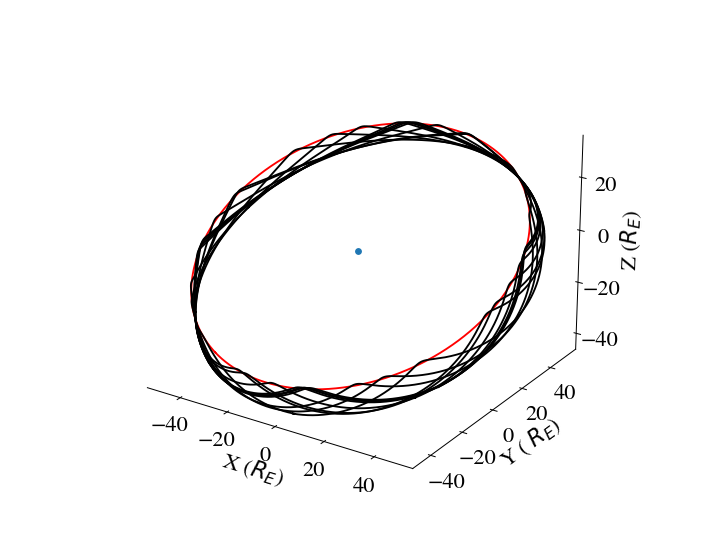}
        \caption{Geocentric inertial reference frame. The red line denotes the Moon's orbit over 1 sidereal month}
        \label{fig:nrho_geo}
    \end{subfigure}
    \caption{Selection of 10 orbits belonging to the $L_2$-NRHO family, propagated for 1 lunar sidereal period}
    \label{fig:nrho}
\end{figure}

The mission analysis for the Gateway platform considers a Near-Rectilinear Halo Orbit (NRHO) with a period of approximately 7 days, located around the second collinear Euler–Lagrange point $L_2$ of the Earth–Moon Circular Restricted Three-Body Problem (CR3BP), characterised by the mass parameter $\mu = 0.01215$. The Equations of Motion (EOM) written below are expressed in nondimensional units and in the synodic (rotating) reference frame $Oxyz$, centred at the barycentre of the Earth-Moon system, with $x$-axis pointing the Moon from the Earth, its $z$-axis in the angular momentum direction, and its $y$-axis completing the right-hand set\cite{ross_dynamical_2022}:
\begin{subequations}
 \label{eq:eom_CR3BP}
 \begin{align}
     \Ddot{x} &= 2\Dot{y} + U_x, \\
     \Ddot{y} &= -2\Dot{x} + U_y, \\
     \Ddot{z} &= U_z.
 \end{align}
\end{subequations}
The state vector $\mathbf{x} = [x,y,z,\Dot{x},\Dot{y}, \Dot{z}]^T \in \mathbb{R}^6$ collects, in order, the position and velocity components of the massless particle relative to the barycentre of the system in the non-inertial synodic frame. The quantities $U_x, U_y, U_z$ denote the components of the gradient of the pseudo-potential function $U(x,y,z)$ with respect to the position coordinates:
\begin{equation}
\label{eq:potential}
    U(x,y,z) = \frac{1}{2}(x^2 + y^2) + \frac{1-\mu}{r_1} + \frac{\mu}{r_2} + \frac{1}{2}\mu(1-\mu).
\end{equation}
The scalar distances between the particle and the two primary bodies are respectively \(r_1\) and \(r_2\): 
\begin{equation*}
    r_1 = \sqrt{(x+\mu)^2 + y^2 + z^2}, \ r_2 = \sqrt{(x - (1-\mu))^2 + y^2 + z^2}.
\end{equation*}
Figures \ref{fig:nrho} present a selection of ten trajectories belonging to the $L_2$-NRHO family. These orbits are obtained by numerically integrating the EOM (\ref{eq:eom_CR3BP}) over one orbital period, which varies within approximately $\pm 1$ hour around the nominal 7-day period.
In Figure \ref{fig:nrho_syn}, the trajectories are depicted in the synodic reference frame while in Figure \ref{fig:nrho_geo} the same orbits are represented in Earth-centred inertial reference frame (ECI), denoted $O_{\mathrm{XYZ}}$. The transformation between a state vector $\mathbf{x}$ expressed in the synodic frame and the corresponding state $\mathbf{X}$ in the geocentric inertial frame is given by:
\begin{equation}
\label{eq_syn_to_eci}
    \mathbf{X} = \left[\begin{array}{cc}
        \boldsymbol{\Omega}(t) & \mathbf{0}_{3\times 3} \\
        \mathbf{0}_{3\times 3} & \boldsymbol{\Omega}(t)
    \end{array} \right]\left( \mathbf{Lx} + \mathbf{x}_{\mathrm{C}} - \mathbf{x}_{\mathrm{E}}\right).
\end{equation}
The matrix $\mathbf{L}$ performs the dimensionalisation of the synodic state, using the characteristic length ($LU$) and velocity ($LU/TU$) units. The vector $\mathbf{x}_{\mathrm{C}}$ accounts for the Coriolis contribution associated with the rotating frame, with angular velocity vector $\boldsymbol{\omega}_s = [0,0,\omega_s]^T, \omega_s = {1}/{TU}$, while $\mathbf{x}_{\mathrm{E}}$ represents the Earth's state in the synodic frame:
\begin{equation*}
    \mathbf{x}_{\mathrm{C}} = LU\left[ \begin{array}{c}
        \mathbf{0}_{3} \\ \boldsymbol{\omega}_s\times [x, y, z]^T
    \end{array}\right], \ \mathbf{x}_{\mathrm{E}} = LU \left[ \begin{array}{c}
        [-\mu, 0, 0]^T \\ \boldsymbol{\omega}_s\times [-\mu, 0, 0]^T
    \end{array}\right] .
\end{equation*}
The time-dependent rotation matrix $\boldsymbol{\Omega}(t) \in \mathbb{R}^{3\times3}$ maps vectors from the synodic frame to the ECI frame. It is constructed according to the classical Euler $3$–$1$–$3$ rotation sequence, parametrised by the Moon’s orbital elements evaluated at the reference epoch $t_0$: 
\begin{equation}
    \boldsymbol{\Omega}(t) = \boldsymbol{\Omega}_3(\Omega_0)\boldsymbol{\Omega}_1(i_0)\boldsymbol{\Omega}_3(\omega_0 + \nu_0 + \omega_st).
\end{equation}
The angles $\Omega_0, i_0, \omega_0, \nu_0$ denote, respectively, the Right Ascension of the Ascending Node, the inclination, the argument of perigee, and the true anomaly of the Moon at the reference epoch $t_0$, as obtained from ephemeris data. For ease of reference, the matrices below define the elementary rotations by an angle $\alpha$ about axis $1$ (denoted $\boldsymbol{\Omega}_1$) and axis $3$ (denoted $\boldsymbol{\Omega}_3$). The terms $c$ and $s$ denote, respectively, the trigonometric functions $\cos\alpha, \sin\alpha$.
\begin{equation*}
\boldsymbol{\Omega}_1(\alpha) = \left[ \begin{array}{ccc}
       1  & 0 & 0\\
       0  & c & -s \\
       0 &s & 1
    \end{array}\right], \ \ 
     \boldsymbol{\Omega}_3(\alpha) = \left[ \begin{array}{ccc}
       c  & -s & 0\\
       s  & c & 0 \\
       0 &0 & 1
    \end{array}\right].
\end{equation*}
The times instants $t\geq0$ are defined relative to the reference one $t_0$, i.e. instant $t=0$ corresponds to the the actual (physical) time $t_0$.
\begin{table}[htbp]
	\fontsize{10}{10}\selectfont
    \caption{Reference quantities for the Orbit Determination Problem}
   \label{tab:od_constants}
        \centering 
   \begin{tabular}{c | l l } 
      \hline 
      Quantity    &Value & \\
      \hline 
      $\mu$ & 0.01215 & \\
      $LU$ & 398703.0 & (km) \\
      $TU$ & 382981.0 & (s)\\
      $t_0$ & 10th March, 2026 00h00 UT (JD)& (s) \\
      $\Omega_0$  &355.947 & (deg)\\
      $i_0$ &28.404 & (deg) \\
       $\omega_0$ &74.893 & (deg)\\
         $\nu_0$&172.562 & (deg)\\
      \hline
   \end{tabular}
\end{table}
\\
Finally, Table \ref{tab:od_constants} reports the reference units defined in section and used in the following of the work. 
\subsection{Angles-Only Observables}
\begin{figure}[htb]
	\centering\includegraphics[width=0.8\linewidth]{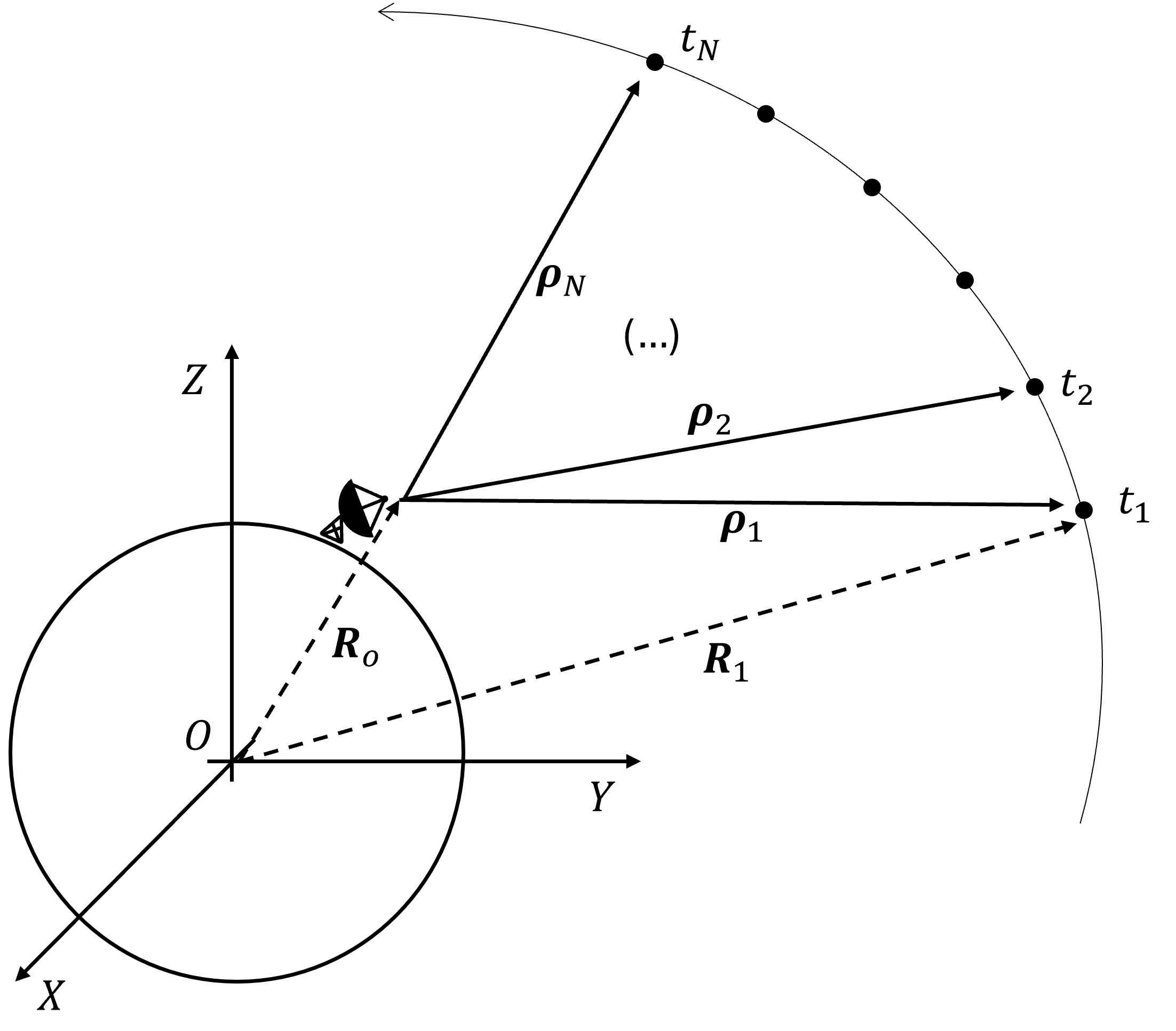}
	\caption{Orbit Determination problem by Angles-Only measurements in ECI reference frame}
	\label{fig:od_1}
\end{figure}

Consider the Orbit Determination problem depicted in Figure \ref{fig:od_1}. An observer is at a position on the Earth's surface, given by vector $\mathbf{R}_o$ with respect to the geocentric frame $O_{XYZ}$. An observation arc is given by the set of observables $\left\{ \mathbf{y}_i \right\}_{i=1}^{N}$, with $N$ the number of observations conducted at successive time instants $t_i$,  $i = 1, 2, \ldots, N$. Those observables are derived from the topocentric position of the observer. The goal of an Orbit Determination problem is to recover the Geocentric state $\mathbf{X}_1 \in \mathbb{R}^6$ of the target object at the beginning of the observation arc, i.e., at instant $t = t_1$. Let be the geocentric states $\mathbf{X}_i$ belonging to the observation arc and their respective position and velocity components, namely, $\mathbf{R}_i$ and $\mathbf{V}_i$:
\begin{equation*}
    \mathbf{X}_i = \left[ \begin{array}{c}
\mathbf{R}_i \\  \mathbf{V}_i
\end{array}\right], \ i = 1,2,\ldots, N.
\end{equation*} The line-of-sight vector $\boldsymbol{\rho}_i$, is defined as the difference between the target's position vector and the observer's position,  in the same geocentric reference frames:
\begin{equation}
\label{eq:relvec}
   \boldsymbol{\rho}_i =  \mathbf{R}_i -\mathbf{R}_o.
\end{equation}
The observables considered in this work are angles-only measurements, namely the azimuth and elevation angles, denoted by $\theta_i$ and $\varphi_i$, respectively: 
\begin{equation*}
    \mathbf{y}_i = \left[\begin{array}{c}
         \theta_i  \\
         \varphi_i
    \end{array}\right], \ i=1\ldots,N.
\end{equation*}These quantities describe the direction of $\boldsymbol{\rho}_i$ in a local topocentric frame centred at the observer. Let $(X_i, Y_i, Z_i)$ denote the components of $\boldsymbol{\rho}_i$ expressed in this local frame. The measurement equations at epoch $t_i$ are then given by: 
\begin{align}
\label{eq:theta}
     \theta_i &= \arctan\left( \frac{Y_i}{X_i}\right) + \varepsilon_{\theta},  \\
     \label{eq:phi}
     \varphi_i &= \arcsin{\left( \frac{Z_i}{||\boldsymbol{\rho}_i||_2}\right) + \varepsilon_{\varphi} }.
\end{align}
 The quantities $\varepsilon_{\theta}$ and $\varepsilon_{\varphi}$ denote the sensor-dependent measurement errors, here modeled as Gaussian noise with standard deviations $\sigma_{\theta}$ and $\sigma_{\varphi}$, respectively. Their values are reported in Table \ref{tab:od_antenna}, together with the observer’s information (namely, latitude, longitude, and altitude). The observer is the Cebreros Ground Station (near Madrid, Spain), operated by the European Space Agency (ESA).
\begin{table}[htbp]
	\fontsize{10}{10}\selectfont
    \caption{Information on the observer's location and features}
   \label{tab:od_antenna}
        \centering 
   \begin{tabular}{c | l l } 
      \hline 
      Quantity    &Value & \\
      \hline 
       Latitude & 40.4528&(deg)\\
        Longitude &4.3675  & (deg) \\
       Altitude &  794 & (m)\\
      $\sigma_{\theta}$& 5e-5 & (rad) \\
      $\sigma_{\varphi}$  &5e-5 & (rad)\\
      \hline
   \end{tabular}
\end{table}

\subsection{Nonlinear Least-Squares State Estimation}
A standard approach to state estimation in the cislunar environment is Nonlinear Least-Squares Estimation (NLSE) based on a Gauss–Newton iterative scheme. The method minimises the discrepancy between measured and predicted observables over an observation arc. In the following, only a concise description is provided; the reader is referred to the relevant literature for a comprehensive treatment. Let
\begin{equation}
\boldsymbol{\sigma}_i =
\left[
\begin{array}{c}
\arctan\left(\dfrac{\sin(\theta_i - \hat{\theta}_i)}{\cos(\theta_i - \hat{\theta}_i)}\right) \\
\varphi_i - \hat{\varphi}_i
\end{array}
\right]
\end{equation} be the residual vector at observation time $t_i$, defined as the difference between the true observables $(\theta_i,\varphi_i)$ and the estimated observables $(\hat{\theta}_i,\hat{\varphi}_i)$. The azimuth residual is expressed using the $\arctan(\sin/\cos)$ formulation in order to properly account for angular periodicity and avoid discontinuities at $\pm\pi$. The estimated observables are obtained by applying the noiseless measurement model of Equations (\ref{eq:theta})-(\ref{eq:phi}) from the observer position at the instant $t_i$ to vector ${\boldsymbol{\rho}}_i$, this last computed with Equation (\ref{eq:relvec}) after numerically integrating the EOM (\ref{eq:eom_CR3BP}) from the estimated initial state $\hat{\mathbf{X}}_1$, expressed in synodic coordinates by appropriately inverting the transformation given in Equation (\ref{eq_syn_to_eci}). The Gauss–Newton update for the initial state estimate at iteration $j$ is derived in the following, with $\boldsymbol{\sigma} \in \mathbb{R}^{2N}$ denoting global residual vector, obtained by stacking all the residuals at any given instant:

\begin{equation}\label{eq:newton}
\hat{\mathbf{X}}_1^{j} = 
\hat{\mathbf{X}}_1^{j-1} - 
\lambda \mathbf{J}^{\dagger}
\boldsymbol{\sigma}^{j-1},
\qquad
j = 2,\ldots,N_{\max}.
\end{equation}
Matrix $\mathbf{J}$ denotes the Jacobian matrix of the global residual vector to the adimensional initial state \begin{equation*}
    \mathbf{J} =\frac{\partial \boldsymbol{\sigma}^{j}}
{\partial \hat{\mathbf{x}}_1^{j-1}} \in \mathbb{R}^{6\times2N},   
\end{equation*}
and  $(\cdot)^{\dagger}$ the inverse matrix;  $\lambda = 1$ is a damping coefficient introduced to improve convergence robustness. Note that for computational convenience the adimensional version of the initial state is used. 
The inverse of the Jacobian is evaluated using a truncated Singular Value Decomposition (SVD). If
\begin{equation}
\mathbf{J} = \mathbf{U}\mathbf{S}\mathbf{V}
\end{equation} is the SVD of the Jacobian, only singular values greater than a prescribed threshold (here $\num{1e-4}$) are retained. The pseudoinverse is then constructed as 
\begin{equation}
\mathbf{J}^{\dagger} = 
\mathbf{V}^{\mathrm{T}}
\mathbf{S}^{*}
\mathbf{U}^{\mathrm{T}},
\end{equation} 
where $\mathbf{S}^{*}$ contains the reciprocals of the retained singular values and zeros elsewhere; $(\cdot)^{\mathrm{T}}$ denotes the transpose matrix. The iterative process described in Equation (\ref{eq:newton}) is repeated until either a maximum number of iterations $N_{\mathrm{max}} = 100$ is reached or convergence is achieved. Convergence is assessed by monitoring the norm of the difference between successive estimates of the initial state, according to
\[
\left\| \hat{\mathbf{x}}_1^{\,j} - \hat{\mathbf{x}}_1^{\,j-1} \right\| \leq \num{1e-9}.
\]

\subsection{Dataset}
\begin{figure}[htb]
	\centering\includegraphics[width=0.9\linewidth]{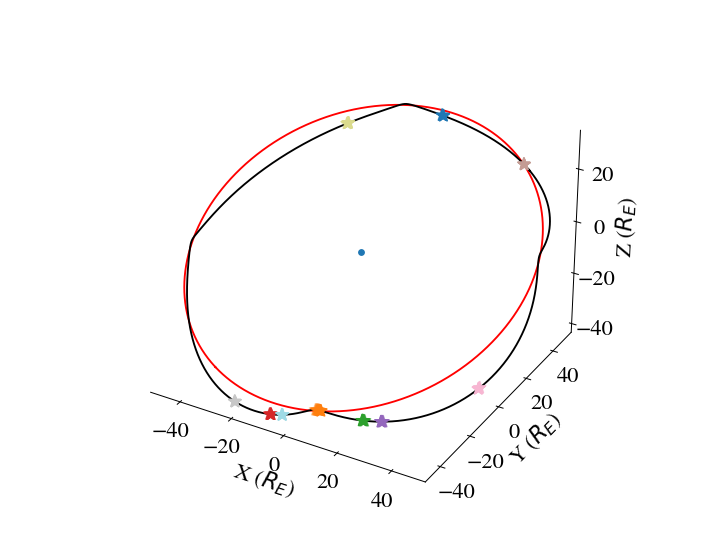}
	\caption{ECI frame: example a the target orbit (black) with a selection of 10 observation arcs. In red: Moon's orbit}
	\label{fig:od_observation_arcs}
\end{figure}

A dataset of synthetic measurement arcs is generated starting from the target orbits expressed in the ECI frame, as shown in Figure \ref{fig:nrho_geo}. More specifically, for each of the 10 selected orbits, 100 observation arcs are constructed, each consisting of 10 observations computed using Equation (\ref{eq:relvec}). For every arc, both the initial observation time and the time interval between consecutive observations are randomly selected in order to produce a diverse set of observation conditions. Observation cadences include measurements taken every 5 minutes (corresponding to the integration time step), every 10 minutes, every 15 minutes, and sequences alternating between 5 and 10 minutes. As a result, the total duration of the observation arcs ranges from a minimum of 50 min to a maximum of 150 min. To illustrate the observation scenario, Figure \ref{fig:od_observation_arcs} shows the target orbit (in black) together with the Moon’s orbit in the ECI frame. A representative subset of 10 observation arcs is also displayed as star markers, with colours distinguishing the different arcs. This visualization highlights the intrinsic difficulty of the orbit determination problem, which is characterised by relatively short observation intervals that challenge accurate estimation of the orbital state.
\\
The dataset of synthetic observations is then constructed by stacking the time instants together with the angles-only observables and applying the measurement model of Equations (\ref{eq:theta})–(\ref{eq:phi}). For each observation arc, 10 measurement realisations are generated by sampling from the noise distribution. We denote by $\mathbf{Y} \in \mathbb{R}^{d \times 3N}$ the resulting dataset of topocentric observations, while the corresponding initial states are stored in $\Bar{\mathbf{X}}_1 \in \mathbb{R}^{d \times 6}$. Thus, for each row index $k \leq d$: 
\begin{equation*}
    \mathbf{Y}_k = [\ldots, t_i, \theta_i, \varphi_i, \ldots], \ \Bar{\mathbf{X}}_{1,k} = \mathbf{X}_1, \ i=1,\ldots, N, \ \ k = 1, \ldots, d.
\end{equation*}
Given the procedure described above, the original dataset consists of 10 000 measurement arcs. However, to represent a more realistic observation scenario, only feasible measurements are retained. Specifically, an observation arc is considered valid only if the target elevation is positive at every observation within the arc:
\begin{equation*}
    \varphi_i \geq 0, i=1, \ldots, N.
\end{equation*}
The resulting dataset has size $d = 4119$ and is subsequently shuffled and split into training ($n_{\mathrm{train}} = $3119 samples), validation ($n_{\mathrm{val}} = $500 samples), and test ($n_{\mathrm{test}} = $500 samples) subsets.
\section{Generative Astrodynamics}
In this Section the generative modelling of the Orbit Determination problem is presented, introducing first the architecture used and then detailing the implementation of the physics-informed approach.
\subsection{Conditional Normalizing Flow Models}
Although the Orbit Determination problem is a well-known inverse problem, it is also meaningful to interpret it as the task of learning a probability distribution from a finite set of samples, that is, as a \textit{generative} modelling problem \cite{kobyzev2021}. This perspective enables a variety of advantages and applications, most notably uncertainty quantification of the estimated target state. More in detail, the target orbit initial state $\Bar{\mathbf{X}}_{1,k}$ can be intended to be the realisation of an unknown probability distribution conditioned to the set of observations $p(\Bar{\mathbf{X}}_{1,k}|\mathbf{Y}_k)$, i.e., $\Bar{\mathbf{X}}_{1,k} \sim p(\Bar{\mathbf{X}}_{1,k}|\mathbf{Y}_k)$. Normalising flow models provide an exact and tractable formulation of this conditional data likelihood through a sequence of invertible transformations \cite{dinh2015nice, papamakarios2019}. Let be $\mathbf{w} \in \mathbb{R}^{6}$ a random variable with a known and tractable probability density function $p(\mathbf{w})$, and let $f: \mathbb{R}^{6}\rightarrow \mathbb{R}^{6}$ the differentiable and invertible mapping conditioned on the observations:
\begin{equation}
\label{eq:flow}
    \mathbf{w} = f(\Bar{\mathbf{X}}_{1,k}|\mathbf{Y}_k), 
\end{equation}
so that 
\begin{equation}
\label{eq:inverse}
\Bar{\mathbf{X}}_{1,k} = f^{-1}(\mathbf{w}|\mathbf{Y}_k) = g(\mathbf{w}|\mathbf{Y}_k)
\end{equation}
is the inverse transformation. By the change-of-variables formula, the log-likelihood of the target distribution can be written as:
\begin{equation}
\label{eq:logvar}
    \log p(\Bar{\mathbf{X}}_{1,k}|\mathbf{Y}_k)  = \log p(\mathbf{w}) + \log\left| \mathbf{J}(f(\Bar{\mathbf{X}}_{1,k}|\mathbf{Y}_k))\right|, 
\end{equation}
with $\left| \mathbf{J}(f(\Bar{\mathbf{X}}_{1,k}|\mathbf{Y}_k))\right|$ denoting the determinant of the Jacobian matrix $\mathbf{J}(f) \in \mathbb{R}^{6 \times 6}$ of the mapping evaluated at state vectors $\Bar{\mathbf{X}}_{1,k}$. 

This work adopts a Real Non-Volume Preserving (Real-NVP) normalising flow \cite{dinh2017realnvp}, consisting of a composition of $n_a = 16$ affine coupling transformations (layers) $f_j, j=1, \ldots,n_a$,
\begin{equation*}
    f = f_{n_a} \circ f_{n_a -1} \circ \ldots \circ  f_{2} \circ  f_{1},
\end{equation*}
each defined as $\mathbf{y}_j = f_j(\mathbf{x}_j|\mathbf{Y}_k)$:
\begin{equation}
    \mathbf{y}_j = \mathbf{b}_j\odot \mathbf{x}_j 
    + (1-\mathbf{b}_j) \odot\!\left(
        \mathbf{x}_j \odot 
        e^{\,s_j(\mathbf{b}_j \odot \mathbf{x}_j,\mathbf{Y}_k )} 
        + t_j(\mathbf{b}_j \odot \mathbf{x}_j,\mathbf{Y}_k )
    \right),
\end{equation}
where $\mathbf{b}_j \in \mathbb{R}^{6}$ is a binary masking vector alternatively masking the position, and velocity components and $\odot$ denotes the Hadamard (element-wise) product. At each coupling layer, the masked components of the input, $\mathbf{b}_j \odot \mathbf{x}_j$, remain unchanged and serve as conditioning variables. These components are concatenated with the observation vector $\mathbf{Y}_k$ and jointly provided as input to the neural networks defining the scale function $s_j(\cdot,\mathbf{Y}_k)$ and translation function $t_j(\cdot,\mathbf{Y}_k)$. The resulting scale and translation outputs are then applied exclusively to the complementary (unmasked) components $(1-\mathbf{b}_j)\odot \mathbf{x}_j$, while the masked components are passed through the layer without modification. This structure ensures a tractable form of the Jacobian, whose log-determinant is simply written as
\begin{equation*}
    \log |\mathbf{J}(f_j)| 
    = \sum_{i=1}^{6} (1-b_{j,i})\, s_j(\mathbf{b}_j \odot \mathbf{x}_j,\mathbf{Y}_k )_i.
\end{equation*}
The learnable transformation functions in this work are parametrised by feedforward neural networks composed of three fully connected layers with hidden dimension $d_h = 256$ and LeakyReLU activations (negative slope 0.1). To ensure numerical stability and prevent excessive Jacobian determinants during training, the outputs from $s_j$ and $t_j$ on the unmasked dimensions are bounded using a hyperbolic tangent activation and rescaled to maximum amplitudes of 0.5 and 0.1, respectively. The total log-likelihood of data sample $\Bar{\mathbf{X}}_{1,k}$, following from Equation (\ref{eq:logvar}), is then written as
\begin{equation}\label{eq:flow_loss_k}
     \log p(\Bar{\mathbf{X}}_{1,k}|\mathbf{Y}_k) 
     = \log p(\mathbf{w}) 
     + \sum_{j = 1}^{n_a} \log \left| \mathbf{J}(f_j)\right|,
\end{equation}
where the base distribution is chosen as a standard multivariate normal $p(\mathbf{w}) \sim \mathcal{N}(\mathbf{0}, \mathbf{I})$, yielding:
\begin{equation*}
    \log p(\mathbf{w}) 
    = -\frac{1}{2}\mathbf{w}^{T}\mathbf{w} 
      - \frac{6}{2}\log (2\pi).
\end{equation*}
\subsection{Generative Orbit Determination}
\begin{figure}[htb]
	\centering\includegraphics[width=\linewidth]{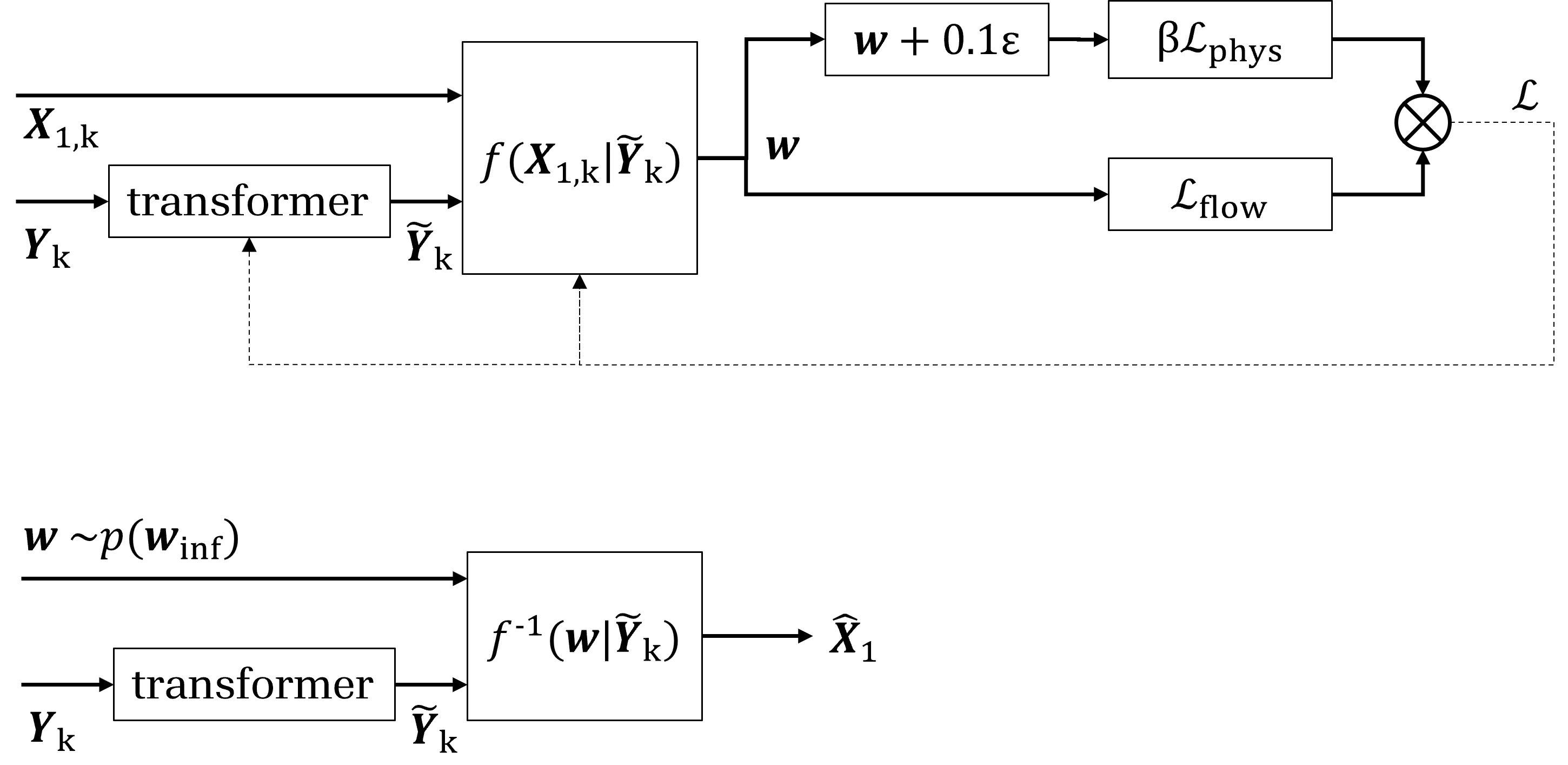}
	\caption{Generative Orbit Determination framework. Top panel: training process, where dashed lines indicate the backpropagation of the loss function. Bottom panel: inference stage}
	\label{fig:gen_od}
\end{figure}

The Orbit Determination architecture combines a Conditional Normalising Flow model with a Transformer-based encoder \cite{vaswani2017attention}. This class of models is particularly effective for a wide range of tasks, including time-series analysis, owing to the attention mechanism that enables the extraction of long-range dependencies within the data. In the present framework, the Transformer processes the observation sequence $\mathbf{Y}_k$ and produces a fixed-size representation $\tilde{\mathbf{Y}}_k \in \mathbb{R}^{256}$. The purpose of this encoding is to obtain an architecture capable of handling observations originating from different observers and locations, while remaining adaptable to other data modalities (for example, pre-processed images). Referring the reader to specialised literature for a detailed description, the Transformer used in this work consists of $n_h = 16$ attention heads and $n_l = 8$ stacked layers.
\\ \\ Figure \ref{fig:gen_od} illustrates the overall Generative Orbit Determination framework. During training, the model learns the conditional mapping associated with the latent base distribution $p(\mathbf{w})$ via the invertible transformations defined in Equation (\ref{eq:flow}). At inference time, samples are drawn from the latent distribution and mapped through the inverse flow conditioned on new observations, thereby producing estimates of the target state at the beginning of the observation arc. In this work, sampling is performed in a local neighbourhood of latent codes obtained from the training data, in a manner analogous to a variational setting. Specifically, perturbed latent samples are generated as
\begin{equation}
    \mathbf{w}_{\mathrm{inf}} = \mathbf{w}_{\mathrm{train}} + 0.1\,\boldsymbol{\varepsilon}, \ \boldsymbol{\varepsilon}\sim\mathcal{N}(\mathbf{0}_6, \mathbf{I}_6),  \ \mathbf{w}_{\mathrm{train}} = f(\Bar{\mathbf{X}}_{1,k}|\mathbf{Y}_k), \ k = 1,\ldots, n_{\mathrm{train}}.
\end{equation}
Here, $\mathbf{w}_{\mathrm{inf}} \in \mathbb{R}^{6}$ denotes the latent variable used at inference time, obtained by adding a small Gaussian perturbation to a latent code $\mathbf{w}_{\mathrm{train}}$ corresponding to a training sample. The perturbed latent vectors are then mapped through the inverse transformation $g(\cdot|\mathbf{Y}_{\mathrm{new}})$ conditioned on the new observations, yielding a set of plausible initial states consistent with the measurement arc. A key advantage of this approach is that multiple samples can be generated for the same observation set. The corresponding outputs constitute a collection of plausible target states consistent with the measurements, enabling uncertainty quantification and improving the robustness of the estimation process.
\subsection{Physics-Informed Loss}
The generative model is trained by minimising a loss function defined as the average per sample of the conditional log-likelihood in Equation (\ref{eq:flow_loss_k}):
\begin{equation}
    \mathcal{L}_{\mathrm{flow}} 
    = - \frac{1}{n_{\mathrm{train}}} 
      \sum_{k=1}^{n_{\mathrm{train}}}
      \log p(\Bar{\mathbf{X}}_{1,k}|\mathbf{Y}_k).
\end{equation}
Optimisation of this objective encourages the model to learn an invertible transformation whose induced latent distribution, when mapped back to the state space, reproduces the empirical distribution of initial states conditioned on the observations. However, likelihood maximisation alone does not guarantee physical consistency. In particular, the flow may generate initial states that are statistically plausible under the learned distribution but dynamically incompatible with the underlying equations of motion, meaning that time propagation of such states does not reproduce the corresponding observation set. To enforce dynamical consistency, an additional physics-informed regularisation term is introduced. The physics-informed loss $\mathcal{L}_{\mathrm{phys}}$ is defined as
\begin{equation}
    \mathcal{L}_{\mathrm{phys}} 
    = \frac{1}{n_{\mathrm{train}}}
      \sum_{k=1}^{n_{\mathrm{train}}}
      \left\| \bar{\boldsymbol{\sigma}}_{k} \right\|^2,
\end{equation}
where $\bar{\boldsymbol{\sigma}}_{k}$ denotes the residual vector associated with sample $k$, obtained by propagating the candidate initial state forward in time according to the equations of motion and comparing the resulting predicted observations with the measured ones. The propagation is performed after transforming the initial state to the synodic frame and subsequently back to the geocentric frame in order to compute the topocentric estimated angular observables.
During training, time integration is carried out using a classical fourth-order Runge–Kutta (RK4) scheme, providing a compromise between numerical accuracy and computational efficiency.
The overall training objective combines the probabilistic flow loss and the physics-informed regularisation through a weighting coefficient $\beta$:
\begin{equation}
    \mathcal{L} 
    = \mathcal{L}_{\mathrm{flow}} 
    + \beta \, \mathcal{L}_{\mathrm{phys}}.
\end{equation}
This composite loss promotes solutions that are both statistically consistent with the training data and physically admissible under the dynamical model, thereby encouraging the learned latent manifold to represent meaningful orbital states rather than purely data-driven interpolations.
\section{Results}
\begin{figure}[htb]
	\centering\includegraphics[width=0.8\linewidth]{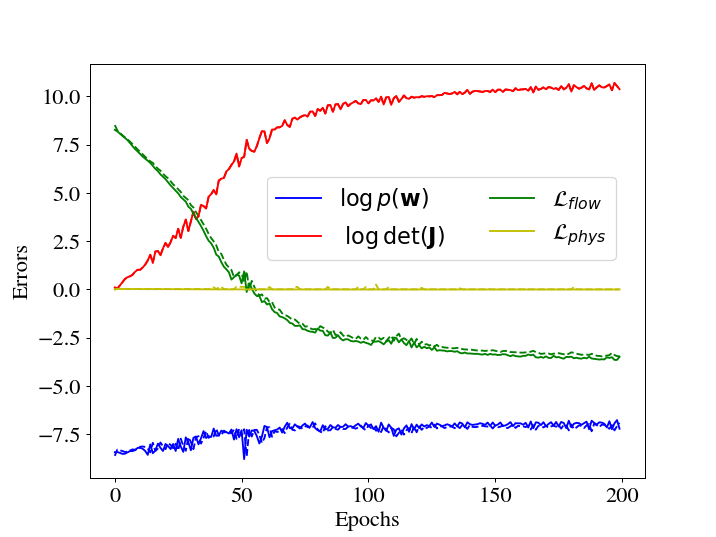}
	\caption{Training dynamics of the physics-informed generative model}
	\label{fig:train_dyn}
\end{figure}

The generative model is trained on the training subset using mini-batches of up to 128 samples. Optimisation is performed for 200 epochs using the Adam optimiser with a learning rate of $\num{1e-4}$. The weighting factor $\beta$ linearly ranges between 0 and 0.5 in 100 epochs and it is then left constant. The evolution of the training process is shown in Figure \ref{fig:train_dyn}, where dashed and solid curves correspond to the training and validation losses, respectively. The blue and red curves represent the two principal components of the flow loss, namely the log-likelihood of the latent variables under the base distribution and the logarithm of the absolute determinant of the Jacobian of the transformation.
It is instructive to compare the learned latent log-likelihood with the value corresponding to an ideal standard Gaussian distribution in six dimensions:
\begin{equation*}
    \log \mathcal{N}(\mathbf{w}) 
    = -\frac{6}{2}\left(1 + \log 2\pi \right) 
    \approx -8.51.
\end{equation*}
The log-likelihood achieved by the flow model is slightly higher (less negative) but remains close to this reference value, indicating that the learned latent distribution approximates a Gaussian while retaining measurable deviations from it. This behaviour is expected, as the true posterior over physically admissible initial states conditioned on the observations is generally non-Gaussian.
These deviations justify the use of a local, variational-style sampling strategy at inference time. By sampling in the neighbourhood of latent codes obtained from the training data, one effectively exploits informative regions of the latent manifold that correspond to dynamically consistent solutions. When mapped back through the inverse flow, such samples yield plausible initial states that are both statistically consistent with the learned distribution and physically compatible with the underlying dynamics.
\begin{table}[htbp]
	\fontsize{10}{10}\selectfont
    \caption{Model training: loss function terms}
   \label{tab:train_od}
        \centering 
   \begin{tabular}{c | c c c c } 
      \hline 
      Subset    &$\log p(\mathbf{w})$ & $\log |\mathbf{J}|$ & $\mathcal{L}_{\mathrm{flow}}$ & $\mathcal{L}_{\mathrm{phys}}$\\
      \hline 
       Training &  -7.0761 &  10.5315& -3.4513 & $\num{8.052e-3}$\\
       Validation & -6.7635 & 10.5315& -3.6475 & $\num{9.090e-4}$ \\
      \hline
   \end{tabular}
\end{table} \\
For ease of reference, Table \ref{tab:train_od} reports a summary of the relevant loss terms for the training and validation sets.
\subsection{Estimation Performance Assessment}
\begin{figure}[htb]
	\centering\includegraphics[width=\linewidth]{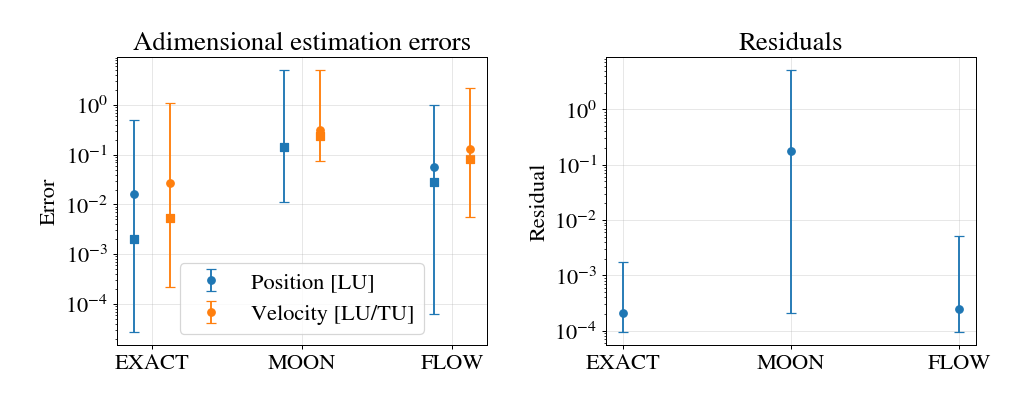}
	\caption{Performance assessment of the estimation process initialised with the exact state, the Moon-based approximation, and the flow model. Left: position and velocity errors; right: post-fit residuals.}
	\label{fig:est_test}
\end{figure}

The primary motivation for adopting a generative approach, as discussed previously, lies in its capability to provide physics- and context-informed initial guesses for standard estimation techniques, such as the Nonlinear Least Squares Estimation (NLSE) method introduced earlier. In this section, the performance of the proposed flow-based model as an initialisation strategy is assessed on the samples of the test subset. For each observation arc, an initial state estimate is generated following the procedure described above and subsequently used to initialise the NLSE algorithm.
\\ To evaluate its effectiveness, the flow-based initialisation is compared against a baseline strategy based on the Moon’s state at the corresponding epoch, computed analytically under the same modelling assumptions adopted throughout this work. This choice is motivated by the proximity of the target orbits to the Moon, which makes such an approximation a plausible practical alternative. In addition, the exact target state is also used as an initial condition for the estimation process, providing a reference benchmark and ensuring a fair comparison across all cases.
The results are summarised in Figure \ref{fig:est_test}. The left panel reports the estimation errors in position and velocity (in normalised units), while the right panel shows the residuals after convergence of the estimation algorithm.
\\
The results in the left panel indicate that the initialisation provided by the flow model consistently outperforms the Moon-based approximation and approaches the performance obtained when initialising from the exact state. This behaviour is observed not only in terms of average performance (mean values, denoted by dots) but also in terms of robustness, as evidenced by the reduced spread and the alignment of median values (squares) across the test set.
The right panel corroborates these findings. The residuals obtained using the flow-based initialisation are of the same order of magnitude as those achieved with the exact initial state, indicating that the estimator converges to physically consistent solutions. In contrast, the Moon-based initialisation yields significantly larger residuals and a wider dispersion, suggesting frequent convergence to local minima rather than to the correct solution.
\subsection{Physics-Informed Loss}
\begin{figure}[htb]
	\centering\includegraphics[width=0.8\linewidth]{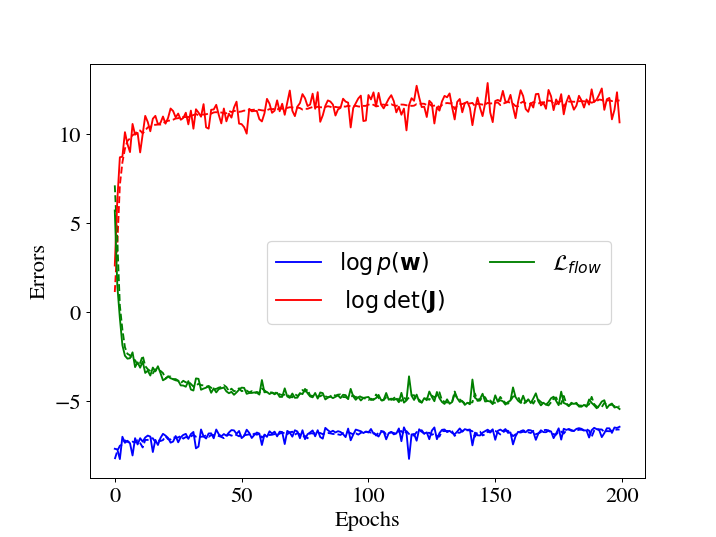}
	\caption{Training dynamics of the physics-agnostic generative model}
	\label{fig:train_dyn_nofis}
\end{figure}

Inspection of the training dynamics reported in Figure \ref{fig:train_dyn} might suggest that the physics-informed loss term $\mathcal{L}_{\mathrm{phys}}$ plays a marginal role during optimisation, as it remains both small in magnitude and nearly constant throughout the training process. This behaviour is, to some extent, expected. Indeed, the normalising flow, by construction, learns an invertible mapping between the data space and the latent space; as a result, it is capable of reconstructing the training samples with high fidelity, implicitly capturing the underlying physical structure of the data through \textit{imitation}, even in the absence of an explicit physics constraint.
\\
However, the generative Orbit Determination problem does not merely require reconstruction of known samples, but rather the ability to generate new, physically consistent states when sampling from the latent distribution conditioned on the observations. This implies that physical admissibility must be enforced not only at the training points, but also in the neighbourhood of the latent manifold where inference is performed.
To further investigate this aspect, the same model was trained under identical conditions but without the physics-informed loss term. The corresponding training dynamics are shown in Figure \ref{fig:train_dyn_nofis}. While the final values of the log-likelihood components are comparable to those obtained in the physics-informed case, the absence of $\mathcal{L}_{\mathrm{phys}}$ results in a less constrained transformation. In particular, the physics-informed loss acts as a regularising mechanism, effectively damping the dynamics of the learned mapping and encouraging the model to preserve the physical consistency of the generated states across the latent space.
\begin{table}[htbp]
	\fontsize{10}{10}\selectfont
    \caption{Model training: loss function terms}
   \label{tab:est_nofis_od}
        \centering 
   \begin{tabular}{c | c c c }
      \hline 
          & Minimum & Mean & Maximum \\
      \hline 
       Position (LU) & \num{3.6225e-5} & 19.4136  & 9675.1388 \\
      Velocity (LU/TU) & \num{2.9750e-3}  & 40.2306 & 20050.2738  \\
      Residual & \num{9.5755e-5} & \num{2.4486e-4} & \num{5.2666e-3} \\
      \hline
   \end{tabular}
\end{table} The impact of this difference becomes more evident when evaluating the estimation performance obtained using the flow-based initialisation. As reported in Table \ref{tab:est_nofis_od}, although the residuals remain relatively small on average, the estimation exhibits significant degradation in terms of robustness. In particular, the large mean and maximum errors in both position and velocity indicate the presence of failure cases, in which the estimator converges to physically inconsistent or suboptimal solutions.
\\ \\
The physics-informed loss acts as a regularisation term on the latent space geometry. By enforcing dynamical consistency during training, it ensures that samples drawn from the posterior distribution are not only statistically plausible but also physically admissible. Consequently, the inclusion of $\mathcal{L}_{\mathrm{phys}}$ is essential to guarantee that the generative model produces reliable initial conditions for downstream estimation tasks, especially in regimes where extrapolation beyond the training data is required.

\section{Conclusion}
This work has demonstrated the application of a generative artificial intelligence framework to the solution of an angles-only orbit determination problem in the cislunar environment, thereby extending the scope of Generative Astrodynamics.
\\ \\
Conditional normalising flows have been shown to effectively approximate the probability distribution of the initial state of an observation arc conditioned on the corresponding measurements. The introduction of a physics-informed loss, based on the numerical integration of the estimated state and the subsequent application of the measurement model, proved instrumental in regularising the learned transformation. In particular, it enforces consistency with the underlying dynamical model in the neighbourhood of the latent samples, thus improving both physical fidelity and generalisation.
\\ \\
At inference time, the proposed approach yields competitive performance when employed as a physics- and context-informed initialisation strategy for classical estimation methods such as Nonlinear Least Squares Estimation (NLSE). In particular, it outperforms initialisation based on the Moon state in terms of both residuals and state estimation errors, while achieving accuracy comparable to that obtained with exact initialisation. This demonstrates the capability of the method to reliably initialise the estimation process, mitigating the risk of convergence to local minima.
\\ \\
Moreover, in contrast to deterministic machine learning approaches, the proposed framework estimates a full conditional probability distribution. This enables the generation of multiple plausible initial states through sampling, supporting more robust estimation pipelines and providing a natural mechanism for uncertainty quantification. The flexibility of the variational sampling process further allows the representation of complex and potentially multi-modal distributions, opening the possibility of jointly modelling heterogeneous families of orbits within a unified latent space.
\\ \\
Owing to its grounding in both physical principles and orbital context, the proposed initialisation strategy has the potential to be extended to a broader class of orbit determination and space navigation problems, including scenarios with sparse or noisy observations, multi-sensor data fusion, and autonomous onboard navigation. Future work will focus on assessing the performance of the method in more realistic settings, including real observational data, as well as investigating scalability, robustness to model mismatch, and integration within operational orbit determination frameworks.
\newpage

\bibliographystyle{AAS_publication}   
\bibliography{references}   

\end{document}